\documentclass[letterpaper, 10 pt, conference]{IEEEtran}  
\IEEEoverridecommandlockouts                              

\usepackage{graphics} 
\usepackage{epsfig} 
\usepackage{mathptmx} 
\usepackage{times} 
\usepackage{amsmath} 
\usepackage{amssymb}  
\usepackage{units}
\usepackage[caption=false,font=scriptsize,justification=raggedright,singlelinecheck=true]{subfig} 
\usepackage{verbatim}  
\usepackage{nicefrac}		
\usepackage{acronym}		
\usepackage{glossaries}		
\usepackage[usenames,dvipsnames]{xcolor}			
\usepackage{cite} 
\usepackage{url} 
\usepackage[multidot]{grffile} 
\usepackage{multirow}
\makeatletter
\let\NAT@parse\undefined
\makeatother
\usepackage[ colorlinks=true, citecolor=black, linkcolor=black, urlcolor=black,  bookmarks=true, bookmarksnumbered=true, pdftex, plainpages=false, pdfpagelabels]{hyperref} 

\newacronym{PSI}{PSI}{Photonic Systems Integration}
\newacronym{ACFR}{ACFR}{Australian Centre for Field Robotics}
\newacronym{CRIS}{CRIS}{Centre for Robotics and Intelligent Systems}
\newacronym{ACRA}{ACRA}{the Australasian Conference on Robotics and Automation}
\newacronym{ACRV}{ACRV}{Australian Centre for Robotic Vision}
\newacronym{USyd}{USyd}{the University of Sydney}
\newacronym{UQ}{UQ}{the University of Queensland}
\newacronym{QUT}{QUT}{the Queensland University of Technology}
\newacronym{UCSD}{UCSD}{the University of California, San Diego}
\newacronym{ANU}{ANU}{Australia National University}
\newacronym{IMOS}{IMOS}{the Integrated Marine Observation System}
\newacronym{URI}{URI}{the University of Rhode Island}
\newacronym{WHOI}{WHOI}{Woods Hole Oceanographic Institution}

\newacronym{NSF}{NSF}{National Science Foundation}
\newacronym{LIEF}{LIEF}{Linkage Infrastructure, Equipment and Facilities}

\newacronym{ICCP}{ICCP}{the International Conference on Computational Photography}
\newacronym{CVPR}{CVPR}{Computer Vision and Pattern Recognition}
\newacronym{TIP}{TIP}{Transactions on Image Processing}
\newacronym{TSP}{TSP}{Transactions on Signal Processing}
\newacronym{JFR}{JFR}{the Journal of Field Robotics}
\newacronym{ISCAS}{ISCAS}{International Symposium on Circuits and Systems}
\newacronym{TOG}{TOG}{Transactions on Graphics}
\newacronym{ICRA}{ICRA}{International Conference on Robotics and Automation}
\newacronym{IROS}{IROS}{Intelligent Robots and Systems}
\newacronym{RA-L}{RA-L}{Robotics and Automation Letters}

\newacronym{AUV}{AUV}{autonomous underwater vehicle}
\newacronym{UAV}{UAV}{unmanned aerial vehicle}
\newacronym{USV}{USV}{unmanned surface vehicle}
\newacronym{UGV}{UGV}{unmanned ground vehicle}
\newacronym{GPS}{GPS}{global positioning system}
\newacronym{SLAM}{SLAM}{simultaneous localisation and mapping}
\newacronym{SfM}{SfM}{structure from motion}
\newacronym{AR}{AR}{augmented reality}
\newacronym{VR}{VR}{virtual reality}
\newacronym{MR}{MR}{mixed reality}
\newacronym[plural=CNNs,firstplural=convolutional neural networks (CNNs)]{CNN}{CNN}{convolutional neural network}
\newacronym[plural=IMUs,firstplural=inertial measurement unit (IMUs)]{IMU}{IMU}{inertial measurement unit}
\newacronym{TOF}{TOF}{time of flight}

\newacronym{MDSP}{MDSP}{multi-dimensional signal processing}
\newacronym{DOF}{DOF}{degree-of-freedom}
\newacronym{RMS}{RMS}{root mean square}
\newacronym{RMSE}{RMSE}{root mean squared error}
\newacronym{SNR}{SNR}{signal-to-noise ratio}
\newacronym{CNR}{CNR}{contrast-to-noise ratio}
\newacronym{PCA}{PCA}{principal component analysis}
\newacronym{MSE}{MSE}{mean squared error}

\newacronym{FIR}{FIR}{finite impulse response}
\newacronym{IIR}{IIR}{infinite impulse response}
\newacronym{DFT}{DFT}{discrete Fourier transform}
\newacronym{FFT}{FFT}{fast Fourier transform}
\newacronym{PSNR}{PSNR}{peak signal-to-noise ratio}
\newacronym{FPGA}{FPGA}{field programmable gate array}
\newacronym{GPU}{GPU}{graphics processing unit}
\newacronym{ASIC}{ASIC}{application-specific integrated circuit}
\newacronym{BW}{BW}{bandwidth}

\newacronym{PSF}{PSF}{point spread function}
\newacronym{SPAD}{SPAD}{single-photon avalanche diode}
\newacronym{FOV}{FOV}{field of view}
\newacronym{BRDF}{BRDF}{bidirectional reflectance distribution function}
\newacronym{FWHM}{FWHM}{full width at half maximum}
\newacronym{LF}{LF}{light field}
\newacronym{2pp}{2pp}{two-plane parameterization}
\newacronym{MLA}{MLA}{microlens array}

\newacronym{RANSAC}{RANSAC}{random sampling and consensus}
\newacronym{DoG}{DoG}{difference of Gaussian}
\newacronym{SIFT}{SIFT}{scale invariant feature transform}

\newacronym{ROS}{ROS}{Robotics Operating System}
\newacronym[plural=EPIs,firstplural=epipolar plane images (EPIs)]{EPI}{EPI}{epipolar plane image}
\newacronym{ReLU}{ReLU}{Rectified Linear Unit}

\title{\LARGE \bf
Unsupervised Learning of Depth Estimation and Visual\\Odometry for Sparse Light Field Cameras
}

\author{S. Tejaswi Digumarti$^{\ast\ \dagger,\S}$, Joseph Daniel$^{\ast\ \dagger}$, Ahalya Ravendran$^{\dagger,\S}$, Donald G. Dansereau$^{\dagger,\S}$
\thanks{$^{*}$ Authors contributed equally to the work presented.}
\thanks{$^{\dagger}$Tejaswi Digumarti, Joseph Daniel, Ahalya Ravendran and Donald Dansereau are with the School of Aerospace, Mechanical and Mechatronic Engineering, The University of Sydney, 2006 NSW, Australia.
        {\tt\small joseph.daniel.sydney.edu.au, arav3215@uni.sydney.edu.au}}
\thanks{$^{\S}$Tejaswi Digumarti, Ahalya Ravendran and Donald Dansereau are with the Sydney Institute for Robotics and Intelligent Systems, 2006 NSW, Australia.
        {\tt\small tejaswi.digumarti, donald.dansereau@sydney.edu.au}}%
}

\newcommand{\myquat}[1]{\bar{#1}}

\newcommand{\q}{\myquat{q}}

\newcommand{\Cq}[2]{\boldsymbol{C}(\q)}

\newcommand{\Equation}[1]{(#1)}
\newcommand{\Figure}[1]{Fig. #1}

\newcommand{\Table}[1]{Tab. #1}
\newcommand{\Section}[1]{Sec. #1}

\begin{document}

\maketitle
\thispagestyle{empty}
\pagestyle{empty}

\begin{abstract}

While an exciting diversity of new imaging devices is emerging that could dramatically improve robotic perception, the challenges of calibrating and interpreting these cameras have limited their uptake in the robotics community. In this work we generalise techniques from unsupervised learning to allow a robot to autonomously interpret new kinds of cameras. We consider emerging sparse light field (LF) cameras, which capture a subset of the 4D LF function describing the set of light rays passing through a plane. We introduce a generalised encoding of sparse LFs that allows unsupervised learning of odometry and depth. We demonstrate the proposed approach outperforming monocular and conventional techniques for dealing with 4D imagery, yielding more accurate odometry and depth maps and delivering these with metric scale.  We anticipate our technique to generalise to a broad class of LF and sparse LF cameras, and to enable unsupervised recalibration for coping with shifts in camera behaviour over the lifetime of a robot. This work represents a first step toward streamlining the integration of new kinds of imaging devices in robotics applications.

\end{abstract}

\section{INTRODUCTION}
\label{sec:intro}

Integrating new imaging devices into robotics applications is a skilled and challenging task.  While an exciting variety of new imaging capabilities is emerging, dealing with calibration, compensating for non-idealities, and interpreting new forms of visual information have historically been time-consuming. This has limited the uptake of new visual sensors in robotics.

Emerging imaging technologies could allow robots to see better in a range of scenarios.  Recent capabilities include imaging around corners, directly observing light propagation, adaptive and long-range depth sensing, and imaging through occluders like rain, snow, and fog using \gls{LF} cameras~\cite{o2018confocal, lindell2018towards, bartels2019agile, dansereau2015linear, bajpayee2018real}. Before these technologies can be used in robotics we must find ways of dealing with their unique characteristics. 

\begin{figure}
	\centering
	\includegraphics[width=\columnwidth]{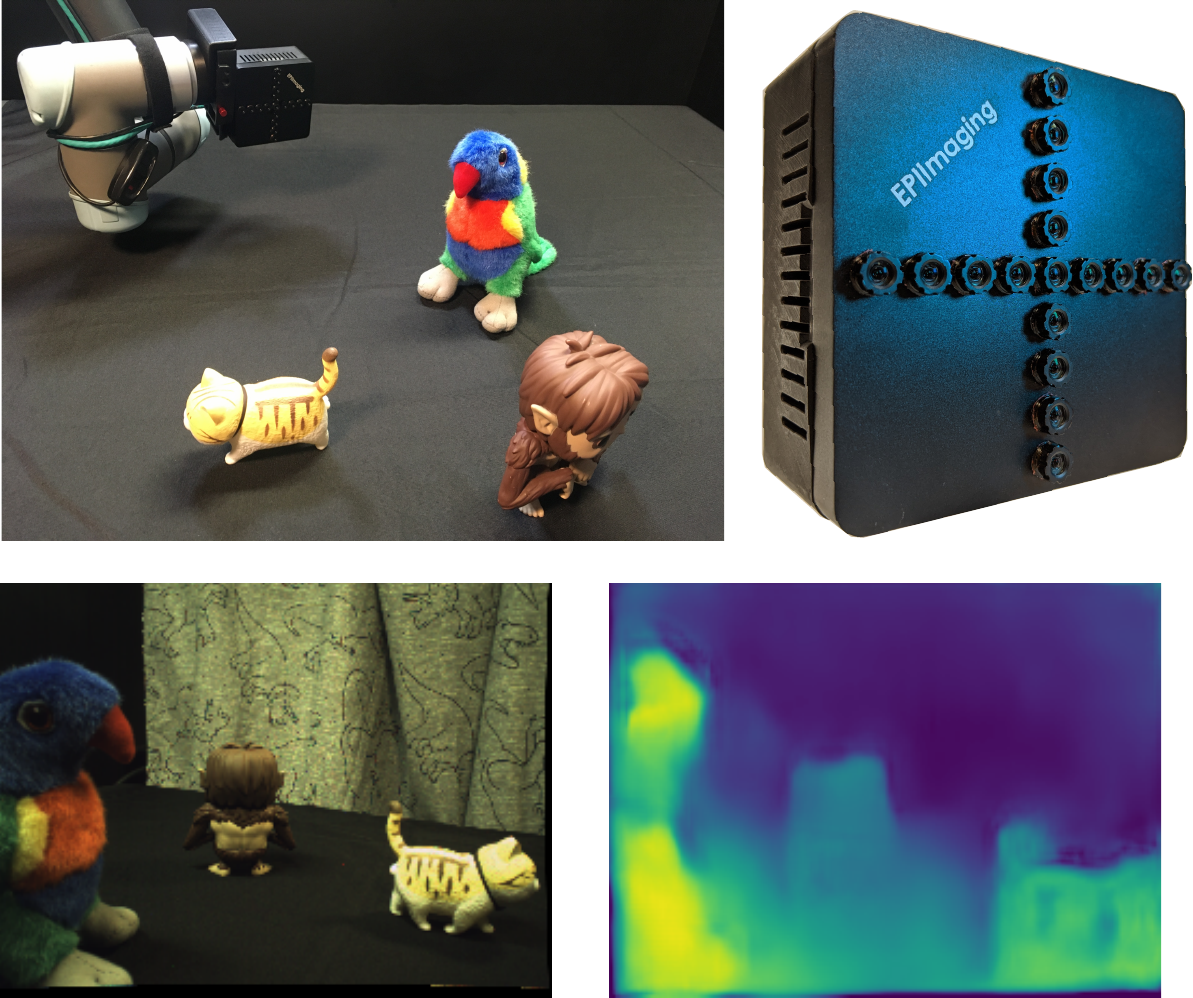}\hfil
	\caption{We propose an unsupervised approach to interpret new imaging devices like the EPIModule from EPIImaging LLC shown here. We propose a novel encoding scheme that benefits from the view diversity of these devices while allowing a broad family of cameras to be used without manual intervention. To demonstrate the technique we learn visual odometry and depth estimation, delivering metric results with greater accuracy and detail than prior approaches.}
	\label{fig_Teaser}
\end{figure}

In this work, we take a step towards the automated interpretation of new cameras by adapting unsupervised learning techniques to deal with sparse \gls{LF} cameras. \gls{LF} cameras in general have been shown to offer improved performance in low light and underwater, and by simplifying conventionally complex tasks like visual odometry and change detection~\cite{bajpayee2018real, dansereau2015linear, dong2013plenoptic, dansereau2016simple}. A full \gls{LF} camera captures a regular grid of views, yielding a 4D image that encodes the behaviour of light in terms of both ray position and direction. A sparse \gls{LF} camera like the one shown in Fig.~\ref{fig_Teaser} captures a subset of these views. This has been shown to offer many of the same advantages as \glspl{LF}~\cite{shi2014light, kalantari2016learning} while using a fraction of the imaging bandwidth. It does however require more sophisticated algorithms for carrying out tasks like visual odometry.

To make sense of these cameras, we leverage recent work in unsupervised learning that shows how using prediction as a training signal one can learn useful tasks without need for costly labelled data~\cite{zhou2017unsupervised, dharmasiri2018eng, zhan2018deepfeature}. We show how to generalise this idea to estimate odometry and depth from sparse \gls{LF} cameras, as a first step toward automating the interpretation of general imaging devices for robotics applications. 

Our key contributions are:
\begin{itemize}
  \item We generalize unsupervised odometry and depth estimation to operate on sparse 4D \glspl{LF};
  \item We introduce an encoding scheme for sparse \glspl{LF} appropriate to odometry and shape estimation, show it outperforming na\"ive \gls{LF} stacking and focal stack approaches, and evaluate using full vs.\ partial \gls{LF} reconstruction as a training signal; and
  \item We demonstrate the proposed methods outperforming the monocular approach, yielding more accurate trajectories and accurate depth maps, with known scale. 
\end{itemize}

To validate our method we mounted an EPIModule from EPIImaging, LLC on a UR5e robotic arm, as shown in Fig.~\ref{fig_Teaser}. We collected 46 trajectories in a variety of indoor scenes, yielding 8298 \glspl{LF}, each with 17 views, and all with accurately known poses as enabled by the robotic arm. We are releasing all data and code along with the paper\footnote{\url{https://roboticimaging.org/Projects/LearnLFOdo/}}.

To evaluate our method we compare against the monocular approach, and more conventional \gls{LF}-based stacking and focal stacking methods. We also compare two prediction modes, one dealing only with the central \gls{LF} view, and the other reconstructing the entire \gls{LF}, drawing on a prior estimate of the inter-camera spacing and camera geometry. We show the proposed methods outperform both monocular and na\"ive \gls{LF}-based  approaches in terms of visual odometry accuracy, photometric warp error, 3D reconstruction accuracy, and qualitative 3D level of detail. 

Our approach captures both the geometric and textural information present in sparse \glspl{LF}, and we expect it to work well for other types of cameras with regular overlapping views. This includes regularly spaced 1D and 2D camera arrays, sparse cameras like the EPIModule, and lenslet-based plenoptic cameras like the Lytro and Raytrix devices. A robot equipped with this capability could swap cameras in and out, requiring only an unsupervised training period to adapt to new imaging hardware.  We anticipate this to be of interest in evaluating new devices and sensor placements in practical applications.


%



\textit{Limitations}:
Although our method is unsupervised and does not require calibration, metric pose and depth require an estimate of the camera layout, including the distances between camera lenses. The module employed in this work conducts onboard distortion correction, but we anticipate the method would work without this feature. Importantly, the learning process can be employed over the lifespan of a robot, allowing it to automatically adapt to shifts in extrinsic or intrinsic camera properties associated with temperature fluctuations, vibrations and fatigue.





\section{RELATED WORK}
\label{sec:related}

\Gls{LF} cameras encode light in a 4D structure that captures light's behaviour in terms of both ray position and direction~\cite{levoy1996lfrendering}. These conventionally parameterise light rays in terms of their points of intersection with two reference planes: an $s,t$ plane, close to the camera array, captures ray position. A second $u,v$ plane placed at an arbitrary distance $D$ and parallel to the first captures ray direction. The combination $s,t,u,v$ uniquely identifies a pixel measured by an \gls{LF} camera, and a ray in world space.

\Glspl{LF} have have been shown to improve imaging performance in challenging conditions~\cite{dansereau2015volumetric, bajpayee2018real} and to simplify a range of tasks, offering effective and sometimes closed-form solutions to both depth estimation and visual odometry~\cite{leistner2019lfdepthcnn, adelson1992single, dong2013plenoptic}.  

Sparse \glspl{LF} capture much of the same information~\cite{shi2014light, kalantari2016learning}, but interpreting their imagery is less obvious, and different configurations offer different tradeoffs in robustness and the algorithmic complexity required for interpretation.  We take the sparse \gls{LF} camera as representative of a class of cameras with overlapping and redundant views. As a step towards demonstrating general autonomous interpretation of newly developed imaging devices, we show how unsupervised learning can be adapted to deal with this class of cameras.  We anticipate the proposed method applies to linear camera arrays, combinations of linear arrays as in the EPIModule shown in Fig.~\ref{fig_Teaser}, and full \glspl{LF} as captured by arrays and lenslet-based cameras.

The use of unsupervised learning has recently emerged as a means of accomplishing complex tasks by cleverly combining prior knowledge of the problem and use of prediction as a feedback mechanism. This has been used to predict camera motion and depth from both monocular and stereo cameras~\cite{zhan2020visual, anisimovskiy2020unsupervised, garg2016unsupervised, zhou2017unsupervised, dharmasiri2018eng, godard2017unsupervised, zhan2018deepfeature}.  These works typically separate the problem into two parts: pose estimation and depth estimation, each handled by a separate network. In the case of stereo cameras some prior knowledge of the camera setup, generally the inter-aperture spacing, is used to obtain metric results.  We draw inspiration from this work and extend it to handle new kinds of cameras.

While prior work has established how to handle monocular and stereo inputs, it is less obvious how one should operate on 4D \glspl{LF}, let alone sparse versions of the same.  Getting this kind of data into a 2D \gls{CNN} is not obvious, and previous work has sliced the 4D \gls{LF} and concatenated the resulting 2D slices into stacks~\cite{heber2017shape}. More sophisticated approaches have sliced in different pairs of dimensions, or interleaved slicing strategies~\cite{wang2016lfcnn}. 

In this work we adopt previous work applying machine learning to \glspl{LF}, and extend them by slicing and concatenating in multiple dimensions, offering the network a mixture of forms of information. We slice in the textural ($u,v$) dimensions, capturing scene appearance, as well as \emph{epipolar} dimensions ($s,u$ and $t,v$), capturing scene geometry. We further build on this by applying a layer of convolutional features to the stacked epipolar slices, similar to the approach used for depth estimation by Shin et al.~\cite{shin2018epinet}. This offers the ability to extract salient geometric features prior to estimating depth and pose. It also has the added advantage of placing the two forms of information, textural and epipolar, in a similar space, facilitating learning. Note that our approach allows the use of conventional 2D \glspl{CNN}, and while some work has generalised to using 3D or even 4D convolutions~\cite{faluvegi2019threedcnn}, this can be more computationally expensive and loses the ability to exploit existing 2D architectures.

To build networks that estimate depth and pose, we leverage prior work that applied a hand-crafted but differentiable warping function to estimate a future image from a prior image, depth map, and relative pose~\cite{zhou2017unsupervised}. As in previous approaches employing stereo imagery~\cite{zhan2018deepfeature}, we consider all input images in the warping process, yielding an estimated \gls{LF} with the same dimensions as the input \glspl{LF}.
This requires a generalisation of the 2D warping function to operate on \glspl{LF}, which we present here along with a comparison to the more naive single-view approach.

\section{METHODS}
\label{sec:methods}


\subsection{Dataset}
\gls{LF} images were collected using an EPIModule from EPIImaging, LLC, mounted on a robotic arm, while executing 46 trajectories. Ground truth poses were recorded for evaluation.
The EPIModule captures images from 17 sub-apertures arranged in a plus sign pattern, as shown in Fig.~\ref{fig_Teaser}. 
The captured images were rectified using off-the-shelf rectification enabled within the module and downsampled to a size of $256 \times 192$ pixels.
A central crop of $224 \times 160$ pixels was then taken from all the images.
The dataset is split into 37 trajectories for training, 6 trajectories for validation and from 3 trajectories for testing.
The test split also contains objects not present during training and validation.

\subsection{Network Architecture}

In keeping with the aim of extending existing unsupervised learning approaches for depth and pose estimation from monocular and stereo images to \glspl{LF}, we draw inspiration from the network architecture presented in~\cite{zhou2017unsupervised}.
The three main components of the network are the following.
\begin{itemize}
	\item A single view depth estimation network that predicts per pixel depth for an input image. We use the encoder-decoder architecture of `DispNet'~\cite{mayer2015dispnet}, with skip connections, as the depth estimation network.
	\item A multi-view pose estimation network (similar to~\cite{zhou2017unsupervised,garg2016unsupervised,bian2019unsupervised}) that takes as input images from two nearby viewpoints and estimates the relative pose of one viewpoint with respect to the other.
	\item A differentiable warp module that couples the depth and pose estimation networks by minimizing a loss based on view synthesis. This loss is the photometric error between a target image and a reference image warped to the viewpoint of the target, using the estimated depth and pose.  
\end{itemize}

\begin{figure}
	\centering
	\includegraphics[width=0.99\columnwidth]{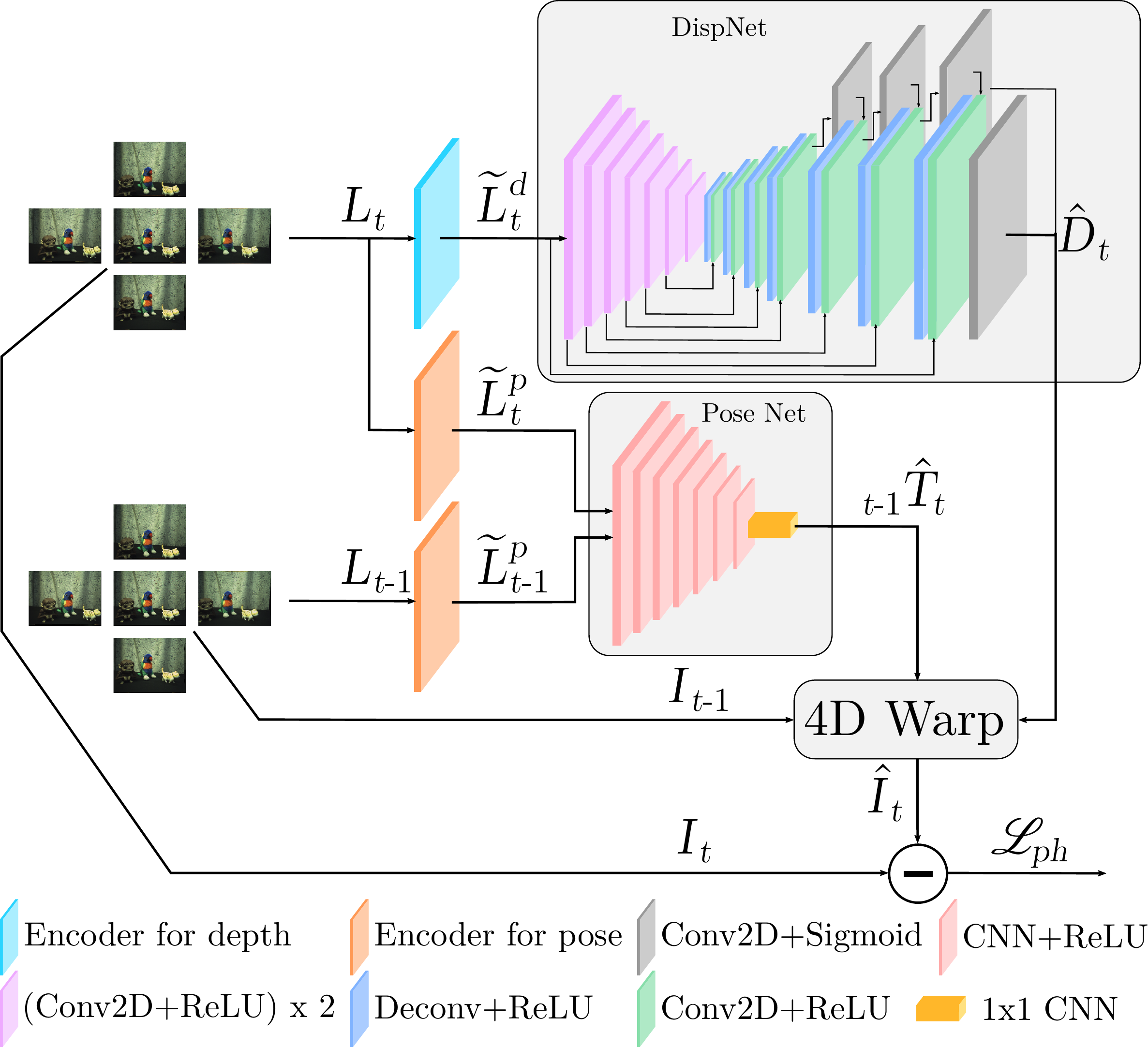}
	\caption{Proposed architecture: Encoders convert each sparse 4D input \gls{LF} $L$ into a form $\tilde{L}^d$ and $\tilde{L}^p$ ingestible by 2D CNNs. From the encoded $\tilde{L}^d_t$, the depth network estimates per-pixel depth $D_{t}$, and from $\tilde{L}^p$ the pose network estimates the pose $_{t-1}\tilde{T}_{t}$ of the camera from time $t-1$ to time $t$. The two networks drive a differentiable warp that predicts an LF $\hat{I}_{t}$ by warping ${I}_{t-1}$ to time $t$. The photometric loss $\mathcal{L}_{ph}$ between the true and estimated LFs drives training of the networks and encoders. We evaluate different encoding schemes, and use of 2D vs.\ 4D warping and photometric loss.}
	\label{fig_TopLevelApproach}
\end{figure}

Generalizing the architecture to sparse \glspl{LF} poses the question of how best 4 dimensional data can be arranged as 2 dimensional slices, so that it forms an informative input to convolutional neural networks that predict depth and pose.
Prior work~\cite{wang2016lfcnn} approached this challenge with the assumption that a complete grid of 2D images is available.
However, the specific configuration of the apertures in the imaging module used in this work (see \Figure{\ref{fig_Teaser}}), prohibits doing so without introducing significant redundant data.
Therefore, we address this challenge by proposing a novel \emph{encoding scheme} that captures both geometric and textural information from the \gls{LF}.
The complete network architecture with the encoders is shown in \Figure{\ref{fig_TopLevelApproach}}.


\subsection{Sparse \gls{LF} Image Encodings}
\label{sec:methods-encodings}

\begin{figure}
	\centering 
	\includegraphics[width=0.99\columnwidth]{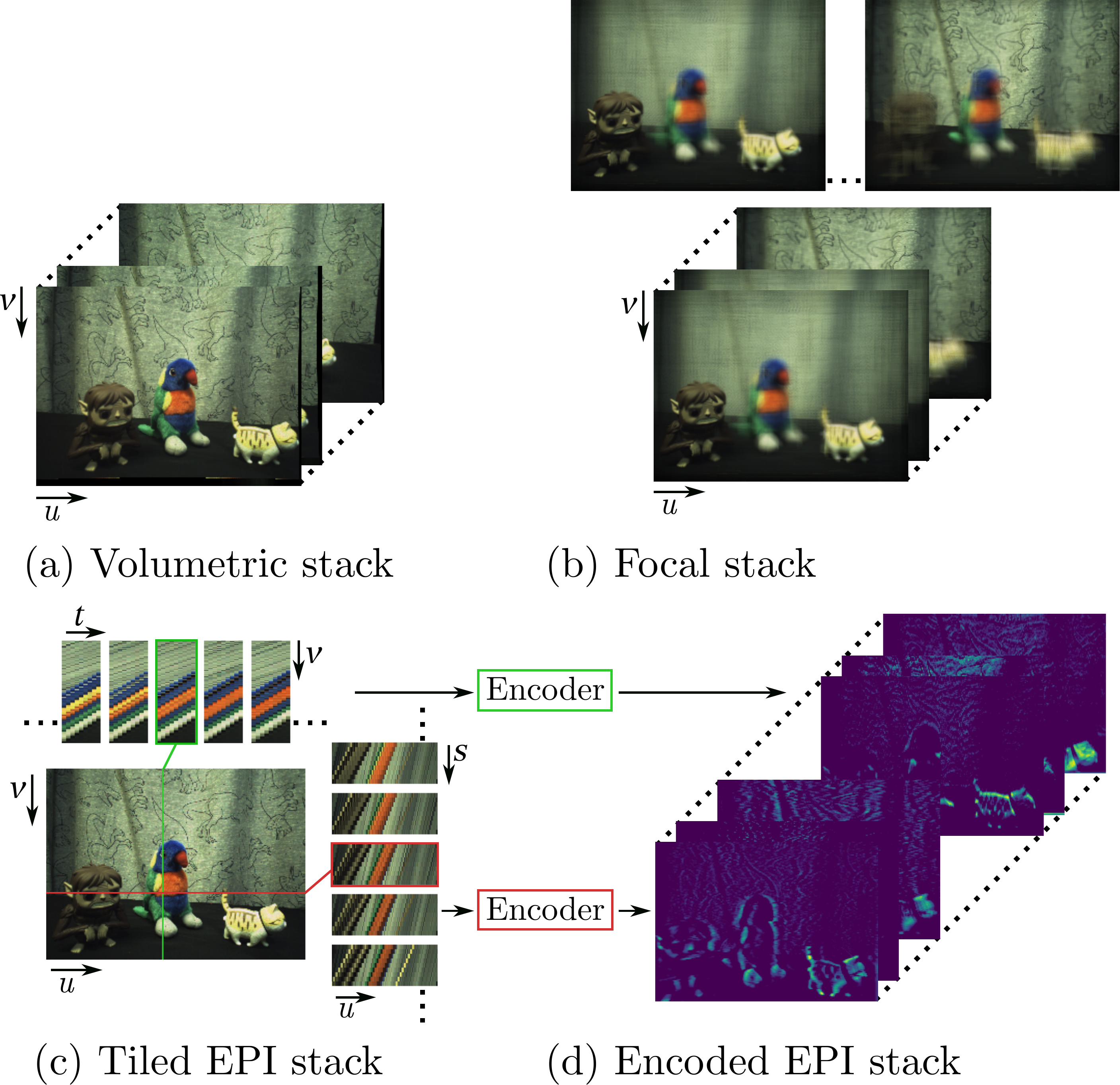}
	\caption{Encodings of a sparse \gls{LF}. (a) Volumetric stacking of images along the colour dimension, (b) Superimposed images in-focus at varying distances from the camera (notice the change in focus from the object in the foreground to the curtain in the background) and the corresponding focal stack, (c) the horizontal and vertical tiling of \glspl{EPI} and (d) the \emph{proposed} encoding scheme where the horizontal and vertical tiled \glspl{EPI} pass through a single CNN layer encoder. The resultant feature maps are stacked to form the proposed encoded \gls{EPI} stack. }
	\label{fig_encodings}
\end{figure}

In order to motivate the choice of the proposed encoding scheme, we first present two approaches of stacking \glspl{LF} as 2D slices in the textural dimension followed by the proposed approach where slicing is performed in the epipolar dimensions.
The three encoding schemes are illustrated in~\Figure{\ref{fig_encodings}}.

\subsubsection{Volumetric Stack} 
A volumetric stack is obtained by stacking images of the sparse \gls{LF} along the $\left(u,v\right)$ direction, i.e. the colour channel dimension. 
This was proposed and evaluated in~\cite{wang2016lfcnn}, for material classification. 
With this encoding, we expect a \gls{CNN} to learn features related to parallax, occlusion and depth. 
For a camera array with $N$ sub-apertures and an image size of $\left(H \times W\right)$ pixels, the volumetric stack has a size of $\left(N \times H \times W\right)$ pixels.

\subsubsection{Focal Stack} 
Superimposing images from each sub-aperture and averaging the intensity at each pixel results in an image where some regions exhibit interference and are `out-of-focus', while other regions are `in-focus' and remain crisp.
Shifting pixels of different sub-apertures by varying amounts prior to superposition, results in images that are in focus at different distances from the camera.
Stacking superimposed images, with different planes of focus, along the colour channel dimension constitutes the focal stack.
As the focal stack encodes depth in the form of interference at each region, we expect a \gls{CNN} to use this information to estimate depth and pose.
However, the trade-off is that aliasing artefacts in the focal stack may affect training. 
For $N$ planes of focus and an image size of $\left(H \times W\right)$ pixels, the focal stack has a size of $\left(N \times H \times W\right)$ pixels.

\subsubsection{Tiled Epipolar Plane Image Stack}
An \gls{EPI}~\cite{bolles1987epipolar} is a slice of the \gls{LF} in the $s, u$ or $t, v$ direction and encodes depth and occlusion information in the slope of the sheared lines in the image.
In order to utilize this information, we propose tiling these images, vertically for the $s, u$ slices and horizontally for the $t, v$ slices, as shown in \Figure{\ref{fig_encodings}c}.
For a camera array with $N$ sub-apertures, such a tiling results in images that are \textit{tall} ($\left(N\cdot W \times H\right)$ pixels) and \textit{wide} ($\left(W \times N\cdot H\right)$ pixels) respectively.

Passing these tiled images as input to the depth and pose estimation networks is not trivial.
Instead of modifying the base architecture to accept these images as inputs, we propose the use of an additional convolutional layer on the tiled \glspl{EPI}, that downsamples them to the shape expected by the networks.
Both the wide and tall tiled \glspl{EPI} are convolved with a kernel of size $\left(N \times N\right)$, but with a horizontal stride of $N$ for the wide tiled \gls{EPI} and a vertical stride of $N$ for the tall tiled \gls{EPI}.
This is followed by a \gls{ReLU} activation layer.
The resulting encoded \glspl{EPI} are stacked along the colour channel to form the \emph{Encoded \gls{EPI} stack}, as shown in \Figure{\ref{fig_encodings}d}.

We input the encoded \gls{EPI} directly to the depth estimation network. 
However, for the pose estimation network, we additionally concatenate images from the volumetric stack.
We hypothesize that by stacking slices in the epipolar dimension with slices in the textural dimension, the pose estimation network can leverage both structural and semantic features from the different input spaces during the learning process.
On the other hand, to improve the ability of the depth network to generalize to unseen objects and not rely on global features in the image, such as the background and the table, we omit the slices in the textural dimension in its input.

\subsection{Loss Formulation}
Given a pair of successive viewpoints, with indices $\left(t-1, t\right)$, from which \glspl{LF} $\left(L_{t-1}, L_t\right)$ are captured, we compute the corresponding encoded \glspl{LF} $\left(\tilde{L}^d_t, \tilde{L}^p_{t-1}, \tilde{L}^p_t\right)$ as described earlier, where the superscripts $d$ and $p$ indicate the specific encoding for the depth and pose estimation networks respectively.
With $\tilde{L}^d_t$ as input, the depth estimation network outputs the pixel-wise depth map\footnote{The network actually estimates inverse depth. For improving readability we omit this technicality.}, $\hat{D}_{c,t}$. We interpret this as the depth map of the central sub-aperture, $c$, of the imaging module corresponding to the \gls{LF} at viewpoint $t$.
Next, given $\tilde{L}^p_{t-1}$ and $\tilde{L}^p_{t}$ as inputs the pose estimation network outputs the relative pose ${}_{c,t-1}\hat{T}_{c,t}$, which we interpret as the relative pose of the central sub-aperture of the imaging module at viewpoint $t$ with respect to the same sub-aperture at viewpoint $t-1$.


Photometric consistency loss can then be computed as

\begin{equation}
	\label{eq:single_warp}
	\mathcal{L}_{ph} = \frac{1}{n}\sum_{i}^n|I_{c,t}(i) - \hat{I}_{c,t}(i)|,
\end{equation}

where $i$ is the index over the pixel coordinates, $n$ is the number of pixels in the image, $\hat{I}_{c,t}$ is the image $I_{c,t-1}$ of the central sub-aperture at viewpoint $t-1$ warped to the viewpoint $t$.

The warped image, $\hat{I}_{c,t}$, is computed by sampling the image $I_{c,t-1}$ with the projected homogeneous pixel coordinates $\mathbf{p}_{c,t-1}$ using differentiable bilinear sampling~\cite{jaderberg2015spatialtransformer}.

The projected homogenous pixel coordinates $\mathbf{p}_{c,t-1}$ are computed from the homogeneous pixel coordinates $\mathbf{p}_{c,t}$ as

\begin{equation}
	\label{eq:single_warp_pixels}
	\mathbf{p}_{c,t-1} \sim K\ {}_{c,t-1}\hat{T}_{c,t}\ \hat{D}_{c,t}\ K^{-1}\ \mathbf{p}_{c,t},
\end{equation}
where K is the matrix of intrinsic parameters of the central sub-aperture.
This is consistent with the approach presented in~\cite{zhou2017unsupervised}.

Additionally, we also employ the multi-scale smoothness loss~\cite{zhou2017unsupervised, garg2016unsupervised, godard2017unsupervised, vijayanarasimhan2017sfm}, to overcome the issue of poor training in low texture regions.
Through empirical evaluation, we also found that replacing the smoothness loss with a loss based on total-variation error~\cite{rudin1992nonlinear} after a few iterations of training helped reduce noisy estimates of depth while preserving edges, especially in low-texture regions.
The total loss was thus a weighted sum of the individual loss terms.

\subsection{Single-warp versus Multi-warp Reconstruction}
We refer to the pipeline described thus far as the \emph{Single-warp} reconstruction pipeline.
We highlight that the single-warp pipeline requires only the knowledge of the intrinsic camera parameters of the sub-apertures and does not rely on the arrangement of the individual sub-apertures within the imaging module.
However, this pipeline suffers from the issue of scale ambiguity.

We address this issue by taking into account the additional information available when imaging a scene using a camera array.
Instead of using the photometric warp to reconstruct the image of a single sub-aperture, the pipeline is modified to reconstruct the \gls{LF}.
Unlike the previous case, the depth estimation network outputs the depth of $M$ sub-apertures instead of a single sub-aperture, while the pose estimation network remains unaltered.
The photometric loss is now computed between the corresponding $M$ sub-apertures across the viewpoints, and the total loss is the mean of the individual losses.

Therefore equations \Equation{\ref{eq:single_warp}} and \Equation{\ref{eq:single_warp_pixels}} can be modified as

\begin{equation}
	\label{eq:multi_warp}
	\mathcal{L}_p = \frac{1}{M\cdot n}\sum_{s \in M}\sum_{i}^{n}|I_{s,t}(i) - \hat{I}_{s,t}(i)|,
\end{equation}
and
\begin{equation}
	\label{eq:multi_warp_pixels}
	p_{s,t-1} \sim K\ {}_{c}{T_{s}}^{-1}\ {}_{c,t-1}\hat{T}_{c,t}\  {}_{c}{T}_{s}\ \hat{D}_{s,t}\ K^{-1}\ p_{s,t}, \forall s \in M
\end{equation}
When $s$ represents the index of a non-central sub-aperture then ${}_{c}{T_{s}}$ is the pose of a non-central sub-aperture relative to the central sub-aperture.
In this work, we assume that this transformation is known, up to a scale factor, and is constant for all the sub-apertures.
However, as stated earlier, this assumption may be relaxed and in turn be predicted using using a photometric consistency constraint similar to \Equation{\ref{eq:single_warp}} imposed between sub-apertures of the same viewpoint.

One can see that this formulation of the photometric error in \Equation{\ref{eq:multi_warp}} penalizes an incorrect estimate of scale because an error in the depth estimate results in a large photometric error for the other sub-apertures.
We call the reconstruction pipeline with this modification the \emph{Multi-warp} reconstruction pipeline.

\section{RESULTS}
\label{sec:results}


\subsection{Implementation Details}
We evaluate three different encodings of the sparse \gls{LF} as described in \Section{\ref{sec:methods-encodings}}.
We consider two variants of the focal stack, one with 5 planes of focus (coarser spacing between the planes) and one with 9 planes of focus (finer spacing between the planes).
We refer to the two configurations as focalstack-5 and focalstack-9 respectively. 
Furthermore, we compare the performance of all the encoding schemes and reconstruction pipelines against the monocular depth and pose estimation approach from~\cite{zhou2017unsupervised}.

When the proposed encoded \gls{EPI} stack is input to the pose estimation network, images from the 4 sub-apertures closest to the central sub-aperture along with that of the central sub-aperture constitute the additional volumetric $\left(u, v\right)$ stack that is concatenated with the \gls{EPI} feature maps.
We also use the same five sub-apertures when computing the photometric error in the multi-warp reconstruction pipeline.

All the networks were trained for 100 epochs, with weights initialized from a Xavier uniform distribution.
The Adam~\cite{Kingma2015AdamAM} optimizer was used during training with a momentum of 0.9 and $\beta$ of 0.999.
We weigh down the smoothness and the total variation loss by a factor of 0.3.

\subsection{Depth Estimation}
\subsubsection{Qualitative Evaluation}
\begin{figure*}
	\centering
	\includegraphics[width=0.8\linewidth]{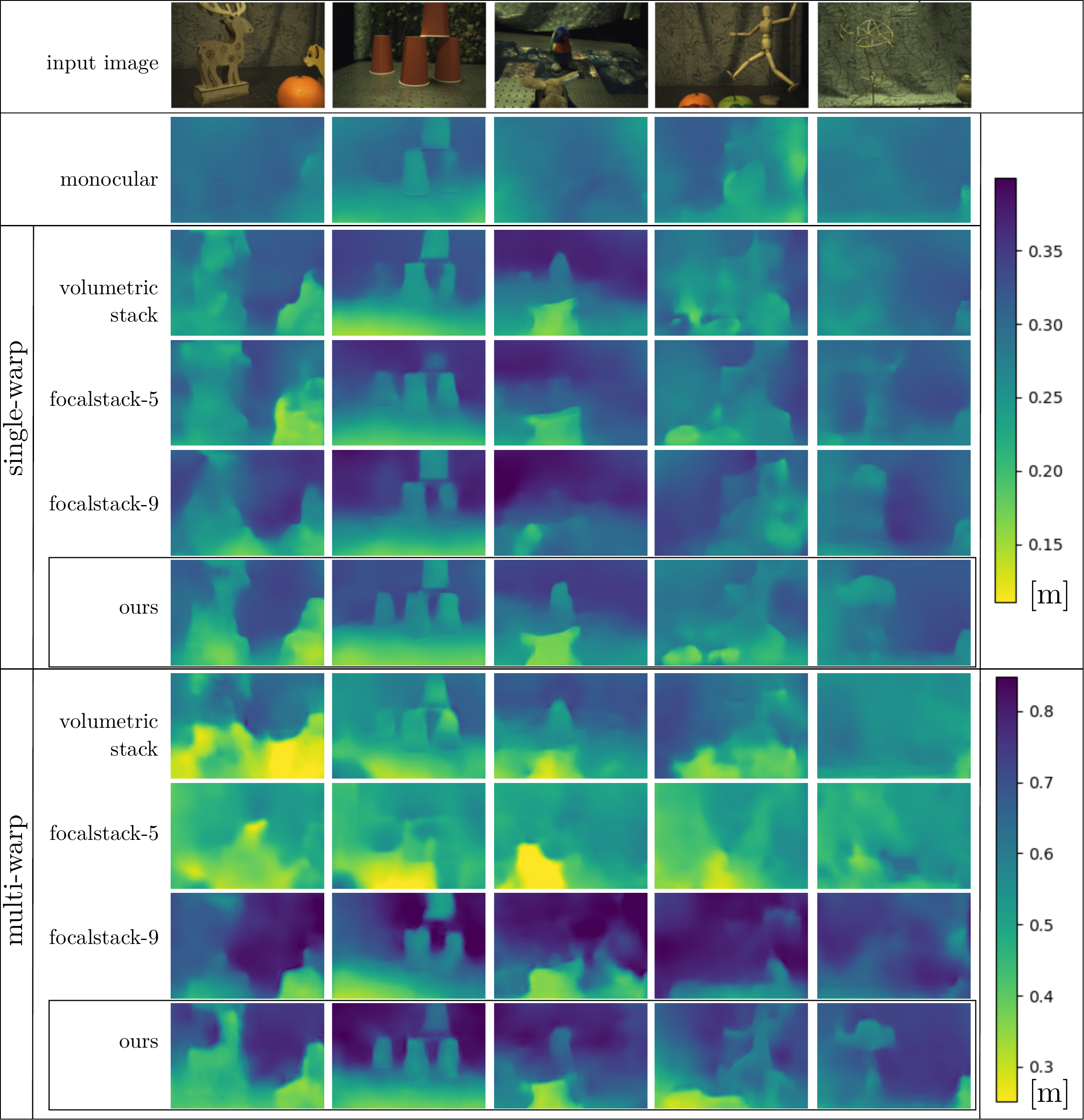}
	\caption{Depth estimates for a few representative images of the dataset for the different input encodings: The network trained with the proposed Encoded Tiled \gls{EPI} stack outperforms the networks with other encoding schemes and monocular depth estimates resulting in better shape estimation and 3D level of detail. It is able to generalize well to previously unseen objects (column 4) and distinguish thin structures (columns 4 and 5), where the other encodings struggle. Note that the colour scales are different between the single-warp and multi-warp pipelines and were chosen for clear visualization.}
	\label{fig:qual_depth}
\end{figure*}

Depth estimates for a few representative images of the test dataset are shown in \Figure{\ref{fig:qual_depth}}.
For both the single-warp and the multi-warp pipelines, the network trained with the proposed encoding outperforms the other encoding schemes, with better shape estimation and greater level of detail in the depth estimates.
The proposed encoding scheme also enables the network to estimate the overall shape of challenging and thin structures (last two columns in \Figure{\ref{fig:qual_depth}}) and at the same time generalize well to previously unseen objects (column 4), while the other methods fail to even detect the object.
This clearly shows the advantage of incorporating both textural and geometric information in the encoding.



\begin{figure}[h]
	\centering
	\includegraphics[width=0.85\linewidth]{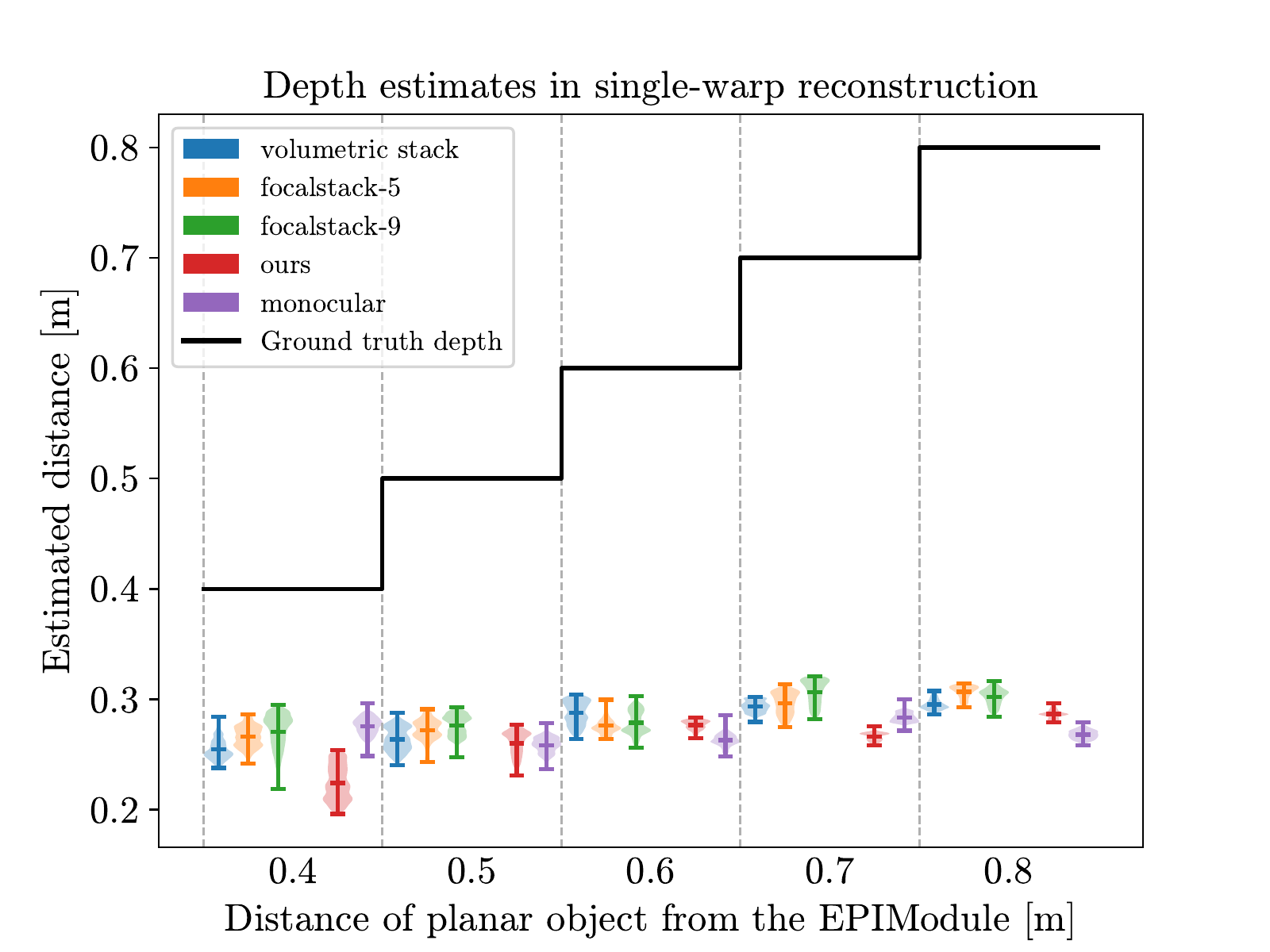}
	\caption{Estimates of depth in the single-warp pipeline for a planar object placed at multiple distances from the imaging module. The single-warp pipeline suffers from scale ambiguity and as a result the estimates are far from the ground truth depth (black).}
	\label{fig:planes_single}
\end{figure}

\begin{figure}[h]
	\centering
	\includegraphics[width=0.85\linewidth]{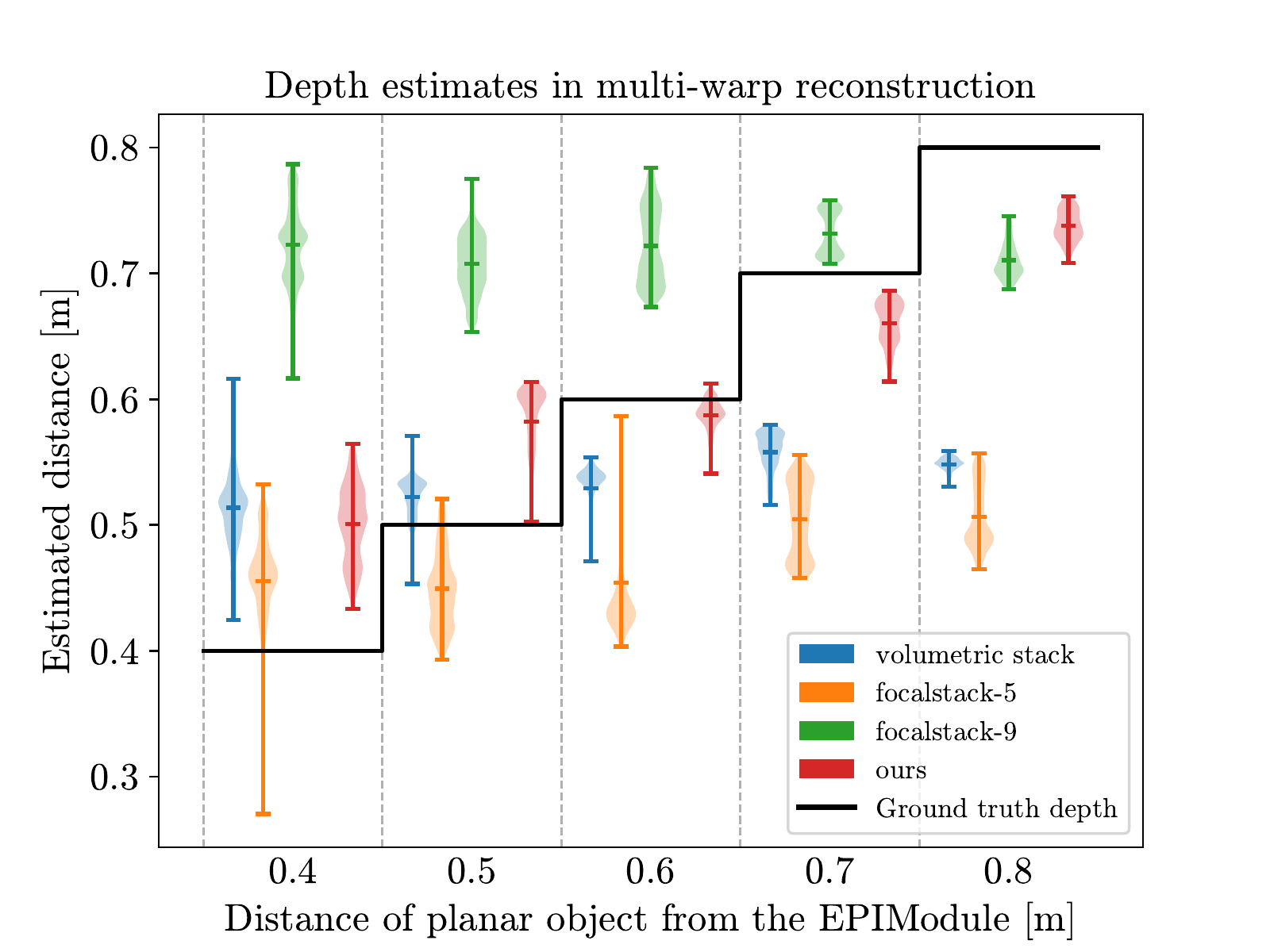}
	\caption{Estimates of depth in the multi-warp pipeline for a planar object placed at multiple distances from the imaging module. The network trained with the proposed encoding scheme (red) is able to estimate depths that are reasonably close to the ground truth depth (black), while with the other encoding schemes the network struggles to learn accurate scale. During training most objects were placed \unit{0.4-0.7}{m} away from the imaging module, and scenes at \unit{0.8}{m} yield reasonable results despite being outside this range.}
	\label{fig:planes_multi}
\end{figure}

\subsubsection{Quantitative Evaluation}

To evaluate the accuracy of the depth estimates, we placed a planar object, with a random texture on it, at multiple known distances away from the EPIModule, fronto-parallel to the imaging planes. 
Depths estimated for each of the encodings was compared against the ground-truth depth. 
These are presented in \Table{\ref{tab:depth-planes}} and illustrated in \Figure{\ref{fig:planes_single}} and \Figure{\ref{fig:planes_multi}}.
We notice that due to the aforementioned ambiguity of scale in the monocular and single-warp reconstruction pipelines, the estimated depth is far from the actual value (see~\Figure{\ref{fig:planes_single}}), despite some networks being able to estimate qualitatively good shape (see~\Figure{\ref{fig:qual_depth}}).
On the other hand while the multi-warp pipeline improves overall scale estimate, only the proposed encoding scheme shows a trend that follows the true depth values, albeit with some error at close and far distances from the module.
Note that during training most scene content was \unit[0.4-0.7]{m} from the imaging module, but the proposed encoding generalizes somewhat beyond this range as seen in~\Figure{\ref{fig:planes_multi}}.

\begin{table*}[h!]
	\caption{Quantitative evaluation of depth estimates: The network with the proposed encoding scheme estimates values close to ground truth in the multi-warp reconstruction pipeline}
	\label{tab:depth-planes}
	\centering
	\begin{center}
		\begin{tabular}{ |c|c|c|c|c|c|c|c|c|c|c|c|c| } 
			\hline
			\multicolumn{2}{|c|}{{\multirow{3}{*}{Method}}} & \multicolumn{10}{|c|}{Distance of planar object from the EPIModule [m]} & \\
			\cline{3-12}
			\multicolumn{2}{|c|}{} & \multicolumn{2}{|c|}{0.4} & \multicolumn{2}{|c|}{0.5} & \multicolumn{2}{|c|}{0.6} & \multicolumn{2}{|c|}{0.7} & \multicolumn{2}{|c|}{0.8} & Overall\\
			\cline{3-12}
			\multicolumn{2}{|c|}{} & Mean & Std.dev & Mean & Std.dev & Mean & Std.dev & Mean & Std.dev & Mean & Std.dev & RMSE [m]\\
			\hline
			\multicolumn{2}{|c|}{Monocular} & 0.275 & 0.008 & 0.258 & 0.007 & 0.263 & 0.006 & 0.283 & 0.004 & 0.268 & 0.004 & 0.359 \\
			\hline
			\multirow{4}{*}{\rotatebox[origin=c]{90}{Single}}
			& volumetric stack & 0.255 & 0.009 & 0.263 & 0.011 & 0.287 & 0.011 & 0.293 & 0.005 & 0.295 & 0.005 & 0.345\\
			\cline{2-12}
			& focalstack-5 & 0.266 & 0.010 & 0.271 & 0.010 & 0.276 & 0.007 & 0.296 & 0.010 & 0.307 & 0.006 & 0.341\\
			\cline{2-12}
			& focalstack-9 & 0.270 & 0.016 & 0.276 & 0.009 & 0.279 & 0.011 & 0.306 & 0.011 & 0.302 & 0.007 & 0.339\\
			\cline{2-12}
			& ours & 0.224 & 0.016 & 0.260 & 0.011 & 0.276 & 0.004 & 0.266 & 0.003 & 0.286 & 0.004 & 0.359\\
			\hline
			\multirow{4}{*}{\rotatebox[origin=c]{90}{Multi}} 
			& volumetric stack & 0.514 & 0.031 & \textbf{0.522} & 0.018 & 0.529 & 0.019 & 0.558 & 0.016 & 0.548 & 0.006 & 0.143\\
			\cline{2-12}
			& focalstack-5 & \textbf{0.455} & 0.040 & 0.449 & 0.030 & 0.454 & 0.044 & 0.505 & 0.029 & 0.507 & 0.025 & 0.174\\
			\cline{2-12}
			& focalstack-9 & 0.723 & 0.031 & 0.707 & 0.024 & 0.722 & 0.030 & \textbf{0.732} & 0.017 & 0.711 & 0.014 & 0.185 \\
			\cline{2-12}
			& ours& 0.501 & 0.030 & 0.582 & 0.027 & \textbf{0.587} & 0.012 & 0.660 & 0.017 & \textbf{0.738} & 0.012 & \textbf{0.067}\\
			\hline

		\end{tabular}
	\end{center}
\end{table*}

\begin{table*}
	\caption{Frame-to-Frame Relative Pose Error: The network with our proposed encoding scheme outperforms other encoding schemes, and monocular pose estimates in terms of both translation and rotation error}
	\label{tab:pose}
	\centering
	\begin{center}
		\begin{tabular}{ |c|c|c|c|c|c|c|c| } 
			\hline
			\multicolumn{2}{|c|}{{\multirow{2}{*}{Method}}} & \multicolumn{3}{|c|}{Relative Pose Error in Translation [m]} & \multicolumn{3}{|c|}{Relative Pose Error in Rotation [deg]}\\
			\cline{3-8}
			\multicolumn{2}{|c|}{} & Mean & Std. dev. & RMSE & Mean & Std. dev. & RMSE \\
			\hline
			\multicolumn{2}{|c|}{Monocular} & 0.029 & 0.016 & 0.033 & 1.522 & 0.969 & 1.808\\
			\hline
			\multirow{4}{*}{\rotatebox[origin=c]{90}{Single}}
			& volumetric stack & 0.028 & 0.015 & 0.032 & 1.453 & 0.880 & 1.703\\
			\cline{2-8}
			& focalstack-5 & 0.030 & 0.015 & 0.033 & 1.439 & 0.883 & 1.693\\
			\cline{2-8}
			& focalstack-9 & 0.030 & 0.016 & 0.034 & 1.452 & 0.912 & 1.716\\
			\cline{2-8}
			& ours & \textbf{0.026} & 0.016 & \textbf{0.031} & \textbf{1.308} & 0.885 & \textbf{1.583}\\
			\hline
			\multirow{5}{*}{\rotatebox[origin=c]{90}{Multi}} & volumetric stack & 0.024 & 0.016 & \textbf{0.029} & 1.366 & 0.802 & 1.585\\
			\cline{2-8}
			& focalstack-5 & 0.031 & 0.016 & 0.035 & 1.457 & 0.779 & 1.653\\
			\cline{2-8}
			& focalstack-9 & 0.035 & 0.017 & 0.039 & 1.585 & 0.868 & 1.807\\
			\cline{2-8}
			& ours& \textbf{0.023} & 0.017 & \textbf{0.029} & \textbf{1.282} & 1.311 & \textbf{1.534}\\
			\hline
		\end{tabular}
	\end{center}
\end{table*}

\subsection{Pose Estimation Results}
We evaluate the performance of the pose estimation network by computing the frame-to-frame Relative Pose Error (RPE)\cite{grupp2017evo} between the estimated relative pose between two camera frames and the ground truth relative pose between the same frames.
The evaluation was performed on the three trajectories of the test set, the results of which are summarized in \Table{\ref{tab:pose}}.
Our proposed encoding scheme outperforms all the other encoding schemes and also monocular pose estimates in terms of both translation and rotation error.
This is true for singe-warp as well as multi-warp reconstruction pipelines.
We further see that due to a better estimate of scale, the multi-warp pipeline has better performance than the single-warp pipeline as was expected.

\section{CONCLUSIONS}
\label{sec:concl}
We have presented an approach for adapting existing techniques developed for traditional cameras to novel imaging devices. We show that by incorporating ideas from plenoptic imaging and unsupervised learning one can successfully estimate depth and odometry from sparse \gls{LF} cameras which outperforms the state-of-the-art monocular reconstruction pipelines. 

We anticipate follow-on work in generalising to irregularly sampled \glspl{LF}, as well as to other modalities like event-based cameras, and multi-modal sensing incorporating inertial measurements. As the approach effectively allows ongoing lifelong calibration, adapting to how cameras change over time, the presented work can be extended to allow an autonomous car to adapt to optical shifts due to thermal warping, vibration, or fatigue, or for an underwater robot to adapt to changes in index of refraction or housing deformation associated with temperature, salinity, or pressure shifts. 


%

\section*{ACKNOWLEDGMENT}

We would like to thank EPIImaging LLC and the University of Sydney Aerospace, Mechanical and Mechatronic Engineering FabLab for their support. We acknowledge the University of Sydney HPC service for providing HPC resources that have contributed to the results reported within this paper.

\addtolength{\textheight}{-12cm}

{\small
	\bibliographystyle{IEEEtran}
	\bibliography{./references}
}

\end{document}